%% file: neurips_foa_2020.tex
\documentclass{article}

     \PassOptionsToPackage{numbers}{natbib}
\usepackage[preprint]{neurips_2020}

\usepackage[utf8]{inputenc} 
\usepackage[T1]{fontenc}    
\usepackage{hyperref}       
\usepackage{url}            
\usepackage{booktabs}       
\usepackage{amsfonts}       
\usepackage{nicefrac}       
\usepackage{microtype}      

\usepackage{graphicx}
\usepackage{caption,subcaption}
\graphicspath{{Images_for_submission/}}
\usepackage{wrapfig}

\usepackage {amsthm}
\theoremstyle{plain}                                                       
\newtheorem{theorem}{Theorem}
\usepackage{natbib}

\title{Wave Propagation of  Visual Stimuli\\in Focus of Attention}

\author{%
  Lapo Faggi$^{1}$, Alessandro Betti$^{2}$, Dario Zanca$^{2}$, Stefano Melacci$^{2}$\thanks{Corresponding author.}, Marco Gori$^{2,3}$ \\
  $^{1}$University of Florence, Florence, Italy \\  
  $^{2}$University of Siena, Siena, Italy \\
  $^{3}$Maasai, Universit\`{e} C\^{o}te d'Azur, Nice, France \\
  \texttt{lapo.faggi@unifi.it}, \texttt{alessandro.betti2@unisi.it}, \texttt{dariozanca@gmail.com}\\
  \texttt{\{mela,marco\}@diism.unisi.it} \\
}

\def\R{\mathbb{R}}
\def\laplace{\Delta}
\def\meno{\medskip\noindent}
\def\H{{\cal H}} 
\def\mint{{\int\mskip -18.6mu - \mskip -6mu}}
\def\div{\mathop{\rm div}\nolimits}

\usepackage{xcolor}

\newif\ifdraft
\draftfalse

\newcommand{\draftA}[1]{\ifdraft{\color{orange}#1}\fi}

\newcommand{\h}{H}
\newcommand{\w}{W}
\newcommand{\dw}{DW}
\newcommand{\Retina}{R}
\newcommand{\drag}{d}
\newcommand{\mass}{m}

\usepackage{amsmath}
\usepackage{amssymb}

\usepackage {amsthm}
\theoremstyle{definition}                                                       
\newtheorem{example}{Example}

\begin{document}

\maketitle

\begin{abstract}
Fast reactions to changes in the surrounding visual environment require
efficient attention mechanisms to reallocate computational resources to most
relevant locations in the visual field. While current computational models
keep improving their predictive ability thanks to the increasing availability
of data, they still struggle approximating the effectiveness and efficiency
exhibited by foveated animals. In this paper, we present a
biologically-plausible computational model of focus of attention that
exhibits spatiotemporal locality and that is very well-suited for parallel and
distributed implementations. Attention emerges as a wave propagation process
originated by visual stimuli corresponding to details and motion
information. The resulting field obeys the principle of ``inhibition of
return'' so as not to get stuck in potential holes.  An accurate
experimentation of the model shows that it achieves top level performance in
scanpath prediction tasks. This can easily be understood at the light of
a theoretical result that we establish in the paper, where we prove that 
as the velocity of wave propagation goes to infinity, the proposed
model reduces to recently proposed state of the art gravitational models of
focus of attention.
\end{abstract}

\section{Introduction}
\input introduction.tex

\section{Gravitational models of FOA and Poisson's Equation}
\label{Poisson:sec}
\input poisson.tex

\section{Visual diffusion and wave propagation}
\input reg.tex

\section{Algorithmic issues and Experimental Results}
\input exp.tex

\section{Conclusions}
\input conclusions.tex

\section*{Broader Impact}
Our work is a foundational study. We believe that there are neither
ethical aspects nor future societal
consequences that should be discussed.

\bibliographystyle{plainnat}
\bibliography{biblio}

%
\clearpage
\appendix
\input supp.tex

\end{document}

%% file: introduction.tex
Visual attention plays a central role in our daily activities. While we are
playing sports, teaching a class or driving a vehicle, the amount of
information our eyes collect is dramatically greater than what  we are able
to process~\cite{allport1989visual,koch2006much}. To work properly, we need a
mechanism that at each time instant only locates the most relevant objects, thus optimizing the computational resources~\cite{ungerleider2000mechanisms}. Human visual attention performs this task so efficiently that, at conscious level, it goes unnoticed.

Attention mechanisms have been the subject of massive investigation also in machines, 
especially whenever they are asked to solve tasks related with human perception such as  video compression, 
where loss of quality is not perceivable by viewers~\cite{itti2004automatic,hadizadeh2013saliency}, 
or caption generation~\cite{liu2017attention,chen2018boosted}. Following the seminal works by Treisman et al.~\cite{treisman1980feature,treisman1969strategies} and Koch and Ullman~\cite{koch1987shifts}, as well as
the first computational implementations~\cite{itti1998model}, over the last three decades scientists have presented numerous attempts to model focus of attention~\cite{borji2012state}. 
The notion of  \textit{saliency map} has been introduced, which consists
of a  spatial map that indicates the probability of focussing on each pixel. 
Under the \textit{centralized saliency map hypothesis}, it has been claimed that shifts in visual attention are then 
generated by a winner-take-all mechanism~\cite{koch1987shifts} to select, at each time step, the most relevant location in space. 

%

Some authors have tried to improve the estimation of scanpaths taking into
account the dynamics of the human visual selection process. Preserving the
centrality of the saliency map, an attempt has been made to introduce
hand-crafted human bias to choose subsequent fixations~\cite{le2015saccadic};
similarly, others have tried to formalize the fact that top-level information
cues continue to increase in importance during visual exploration at the
expense of more perceptive low-level
information~\cite{jiang2016learning}. In~\cite{khosla2007bio} the authors
propose a bio-inspired visual attention module that is based on the pragmatic
choice of identifying proto-objects and learning
a ranking to determine the order in which these proto-objects will be attended. 
All of these approaches still assume the centrality of a saliency map so as to perform a long stack of global computations 
over the entire visual field before establishing the next fixation
point. This is hardly compatible with what is done by the human visual system
that most likely begins computing attention in the early stages of
vision~\cite{briggs2007fast, mcalonan2008guarding} and restricts computation to a small portion 
of the available visual information~\cite{treisman1980feature, schlingensiepen1986importance}.

More recently, Zanca et al. proposed approaches~\cite{zanca2017variational,zanca2019gravitational} that are inspired from physics to model the process of visual attention as a continuous dynamic phenomenon. 
The focus of attention is regarded as particle that is gravitationally
attracted by 
virtual masses originated from details and motion in the visual scene. Details are associated with masses proportional to the magnitude of the gradient of the brightness, while masses due to motion are proportional to the magnitude of the optical flow. This framework is applicable to both images and videos, as long as one considers a static image as a video whose frames are repeated at each time step. Moreover, the model proposed in~\cite{zanca2019gravitational} also implements the inhibition of return mechanism, by monotonically decreasing the saliency  of a given area of the retina that has already been explored in the previous moments. Unlike the previous approaches, the prediction of the focus does not rely on a centralized saliency map, but it acts directly on early representations of basic features organized in spatial maps. 
Besides the advantage in real-time applications, these models make it possible to characterize patterns 
of eye movements (such as \textit{fixations}, \textit{saccades} and \textit{smooth pursuit}) and, despite their simplicity, 
they reach the state of the art in scanpath prediction. However, when looking at these gravitational models
from the biological and computational perspective, one promptly realizes that finding the focus of attention
at a certain time does require the access to all the visual information of the retina to sum up the attraction arising from any virtual mass. Basically, those models are 
not local in space.

This paper proposes a paradigm-shift in the computation of the attraction 
proposed in~\cite{zanca2017variational,zanca2019gravitational}, which is inspired by the 
classic link between global gravitational or electrostatic forces and the associated
Poisson equation on the corresponding potential, that can be regarded as a 
spatially local computational model. Interestingly,  Newtonian gravity yields 
an instantaneous propagation of signals, so as a sudden change in the mass density of a given 
pixel immediately affects the focus of attention, regardless of its location on the retina. 
While the link is intriguing, modeling the focus of attention by the force emerging from 
the static nature of the Poisson potential does not give rise to a truly local 
computational process, since one needs solving the Poisson equation for
each frame. 
This means that such a static model is still missing the temporal propagation that take place
in peripheral vision mechanisms. We show that the temporal dynamics which arise from
 diffusion and wave-based mechanisms are  effective to naturally implement local 
 computation in both time and space. The intuition is that attention is also driven by 
 virtual masses that are far away from the current focus by means of wave-based and/or diffusion
 propagation. We discuss the two different mechanisms of propagation and prove their 
 reduction to gravitational forces as the velocity goes to infinity. The experimental results 
 confirm the intuition that wave-based propagation is more effective in transmitting the 
 information coming from virtual masses than diffusion propagation. However, better 
 results are obtained when properly combining these two propagation mechanisms. 
 Our experimental analysis on scanpaths leads to state of the art results which can clearly 
 be interpreted when considering the reduction to the gravitational model for infinite 
 propagation velocity. The bottom up is that we can reach state of the art results by
 a computational model which is truly local in space and time and that it is clearly
 very well-suited for SIMD hardware implementations.

The paper is organized as follow. In Section 2, we give a brief review of gravitational models of attention~ \cite{zanca2019gravitational} and provide their interpretation in terms of the classic Poisson's equation. In section 3 we propose our model and show that as the velocity of propagation goes to infinity, it reduces to~\cite{zanca2019gravitational}. In section 4, we discuss algorithmic issues along with the experimental setup used to test our model, and illustrate the results on saliency and scanpath prediction. Finally,
some conclusions are given in Section 5.  \\

%% file: poisson.tex
According to~\cite{zanca2019gravitational}, the trajectory of  the focus of attention
$t\in[0,T]\mapsto a(t)\in \R^2$ starting at $a(0)=a_0$ with velocity
$\dot a(0)=a_1$ is the solution of the following Cauchy problem:
\begin{equation}\label{FOA-trajectory}
\begin{cases}
\ddot a(t)+\lambda \dot a(t)+\nabla\varphi_0(a(t),t)=0;\\
a(0)=a_0;\\
\dot a(0)=a_1,
\end{cases}
\end{equation}
where $\lambda>0$ and the scalar function $\varphi^0\colon \R^2\times[0,T]
\to\R$ is
defined as follows:
\begin{equation}\label{grav-potential}
\varphi^0(x,t):={1\over 2\pi}\int_{\R^2}\log{1\over \vert x-y\vert}\mu(y,t)\,
dy.
\end{equation}
Here $\vert\cdot\vert$ is the Euclidean norm in $\R^2$ and $\mu\colon
\Retina\subset\R^2\times [0,T]\to[0,+\infty)$ is the mass distribution at a
certain temporal instant that is present on the retina and is determined by
details and motion:
\begin{equation}
\mu(x,t) = \mu_1(x,t)\left(1-I(x,t)\right)+\mu_2(x,t).
\end{equation} 
In particular, $\mu_1 = \alpha_1 \vert\nabla b \vert$, where $b\colon\R\times
[0,T]\to\R$ is the brightness, while $\mu_2 = \alpha_2 \vert v \vert$, where
$v\colon\R\times [0,T]\to\R$ is the optical flow and $\alpha_1$ and
$\alpha_2$ are positive parameters.  The term
$I(x,t)$ implements the inhibition of
return mechanism, and satisfies 
\begin{align}
	I_t+\beta I = \beta \exp(-\vert x-a(t)\vert^2/2\sigma^2),
\label{IoR-eq}
\end{align}
with $0<\beta<1$.
The crucial observation that will be exploited in the next sections to
develop a fully local theory of focus of attention is the fact that the
potential $\varphi^0$ satisfies the Poisson equation on $\R^2$:
\begin{equation}\label{Poisson-equation}
-\laplace\varphi =\mu,
\end{equation}
where $\laplace$ is the {\it Laplacian} in two dimensions. 
Such result, which is the
two-dimensional analogue of the Poisson equation for the classical
gravitational potential  can be checked by direct calculation. More
generally it follows from these two facts (see for
example~\cite{evans2010partial}):
\begin{enumerate}
\item The function $G(x):=1/(2\pi)\log(1/\vert x\vert)$, defined for $x\in
\R^2$, $x\ne0$, is the fundamental solution of Laplace's equation (i.e.
$-\laplace G=\delta$);
\item to get a solution of the Poisson equation $-\laplace u(x)=f(x)$ in
$\R^2$, when $f$ is regular and compactly supported, it is sufficient to
choose $u$ as the convolution of $G$ with $f$.
\end{enumerate}
Because the ``mass'' density $\mu$ is time dependent, and its temporal
dynamics is synced with the temporal variations of the video,
Eq.~\eqref{Poisson-equation} should in principle be solved for any $t$.
In the next section we will discuss instead how the values
of the potential in a spatial neighbour of $(x,t)$ are exploited to estimate the values of
$\varphi$ at $(x,t+dt)$ by interpreting Eq.~\eqref{Poisson-equation} as
an elliptic limit of a parabolic or hyperbolic equations.


%% file: reg.tex
%
The reformulation of FOA gravitational attraction based on the solution of Poisson's equation 
does require discovering the potential $\varphi(x,t)$ due to the virtual masses
at each frame, thus ignoring any temporal relation. This remark clearly underlines 
also the strong limitation of the solution proposed 
in~\cite{zanca2017variational,zanca2019gravitational}, where the gravitational force is
re-computed at each frame from scratch. 
The main idea behind the reformulation presented in this paper  is
that since we expect that small temporal changes in
the source $\mu$  cause small changes in the solution $\varphi$, then
it is  natural to model the potential $\varphi$ by dynamical
equations which prescribe, for each spatial point $x$, how the solution must
be updated depending on the spatial neighborhood of $x$ at time $t-dt$.  
There is in fact an intriguing connection with electrostatics and 
electrodynamics.  We can introduce an explicit temporal
dynamics in Eq.~\eqref{Poisson-equation}
by introducing  the following two ``regularizations'' 
\begin{equation}
\begin{aligned}
&\h:
\begin{cases}
c^{-1}\varphi_t=\laplace\varphi+\mu& \hbox{in}\quad \R^2\times(0,+\infty);\\
\varphi(x,0)=0,& \hbox{in}\quad \R^2\times\{t=0\},
\end{cases}\\
&\w:
\begin{cases}
c^{-2}\varphi_{tt}=\laplace\varphi+\mu& \hbox{in}\quad
\R^2\times(0,+\infty);\\
\varphi(x,0)=0,\quad \varphi_t(x,0)=0& \hbox{in}\quad \R^2\times\{t=0\},
\end{cases}
\end{aligned}
\label{eq:temp-reg}
\end{equation}
where $c>0$.
Problem $\h$ is a Cauchy problem for the heat equation with source $\mu(x,t)$,
whereas problem $\w$ is a Cauchy problem for a wave equation.
The term $c$ in $\h$ represent the diffusivity constant, whereas
the constant $c$ in problem $\w$ can be regarded as the speed of
propagation of the wave.
The reason why we can consider problem $\h$ and $\w$ as {\em temporal
regularizations} of Eq.~\eqref{Poisson-equation} is due to the following
fundamental result.
\begin{theorem}\label{theorem}
Let $\varphi^0$ be the solution, described in Eq.~\eqref{grav-potential}, of
Eq.~\eqref{Poisson-equation}, then the gradients $\nabla\varphi_H$ and
$\nabla\varphi_W$ of the solutions $\varphi_H$ and $\varphi_W$ to  problems
$\h$ and $\w$ in Eq.~\eqref{eq:temp-reg}  (at least pointwise) converge to
$\nabla\varphi^0$ as $c\to+\infty$.
\end{theorem}
\begin{proof}
See supplementary material.
\end{proof}
Notice that the convergence result expressed by Theorem~\ref{theorem} is
given on the gradients of the potentials and not on the potentials
themselves. 
%
The interpretation of this result is actually quite straightforward. For
problem $\h$ it means that the solution of the heat equation in a substances
with high diffusivity $c$, instantly converges to its stationary
value which is given by Poisson equation~\eqref{Poisson-equation}.
For problem $\w$, Theorem~\ref{theorem} turns out to be the two dimensional
analogue of the infinite-speed-of-light limit in electrodynamics and  in
particular it expresses the fact that the retarded potential
(see~\cite{jackson2007classical}), which in
three spatial dimensions are the solutions of problem $\w$, converges to the
electrostatic potential as the speed of propagation of the wave goes to
infinity ($c\to+\infty$)
\footnote{It is worth mentioning that this regularization result 
was not established in two dimensions.}. 
Although both temporal regularization $\h$ and $\w$ achieve the goal of
transforming the Poisson equation into an initial value problem in time from
which all subsequent states can be evolved from, the different nature of
the two PDE determines, for finite $c$, qualitative differences in the
FOA trajectories computed using Eq~\eqref{FOA-trajectory}. Indeed, since
problem $\h$ models a diffusion process, it tends to smooth out details, that
are instead preserved in the wave-based propagation mechanism $\w$. 
For this reason this paper is mostly
concerned with FOA trajectories obtained from potentials that solve
a wave equation instead of a diffusion equation. Hence, in the remainder 
of the paper, we consider the following generalized version of Eq.~\eqref{eq:temp-reg}
\begin{equation}\label{eq:dumped-wave}
\dw:
\begin{cases}
c^{-2}\mass\varphi_{tt}+c^{-1}\drag\varphi_t=\laplace\varphi+\mu& \hbox{in}\quad
\R^2\times(0,+\infty);\\
\varphi(x,0)=0,\quad \varphi_t(x,0)=0& \hbox{in}\quad \R^2\times\{t=0\},
\end{cases}
\end{equation}
where $\drag>0$ is the drag coefficient and $\mass>0$.
Such equation in one spatial dimension (and without the source term
$\mu$) is known as the telegraph equation (see~\cite{evans2010partial}).
More generally, it describes the propagation of a damped wave.
The FOA model proposed in this paper is based on these $\dw$ equations along with the
inhibition of return equation expressed by~\ref{IoR-eq}.

Clearly, Eq.~\ref{eq:dumped-wave} are local in both space and time, which is 
a fundamental ingredient of biological plausibility. In addition, they are very well-suited
for SIMD hardware implementations. At a first sight, Eq.~\ref{IoR-eq} does not 
possess spatial locality. While this holds true in any computer-based retina, 
in nature, moving eyes rely on the principle that you  can simply 
pre-compute $\exp(-\vert x-a(t)\vert^2/2\sigma^2)$ by an appropriate foveal structure. 
Interestingly, the implementation of moving eyes are the subject of remarkable interest
in robotics  for different reasons (see e.g.~\cite{MRucci-2012}).

%% file: exp.tex
The proposed model is evaluated in saliency and scanpath prediction tasks and
compared with state-of-the-art models. Human eye-tracking datasets exist to
provide well-established benchmarks for the evaluation of computational
models of visual attention. In this work we make use of
CAT2000~\cite{mit-tuebingen-saliency-benchmark,borji2015cat2000}, provided by
the MIT Saliency Team, to evaluate the model in saliency
prediction. The CAT2000 training set includes 2000 input stimuli, grouped
into 20 different semantic categories. Since CAT2000 does not provide
temporal information of human visual explorations, to evaluate the proposed
model in the task of scanpath prediction we use a collection of four smaller
datasets (MIT1003~\cite{Judd}, TORONTO~\cite{bruce2007attention},
KOOTSTRA~\cite{kootstra}, SIENA12~\cite{zanca2018fixatons}) of eye-tracking
data, for a total of 1234 input stimuli. All eye-tracking data have been
collected in free-viewing conditions and subjects were exposed to the
stimulus from 3 up to 5 sec. For each stimulus, we simulate $n=5$
different scanpaths for each of the two models $\h$ and
$\dw$, Eq.~\ref{eq:temp-reg} and Eq.~\ref{eq:dumped-wave}, respectively.

\paragraph{Numerical implementation}
From a computational point of view, we must take into account that our
computations are limited to a finite region of space,
the retina $\Retina\subseteq \R^2$. Thus,
to determine the potential and its time evolution on $\Retina$ we have to
impose additional boundary conditions on $\partial \Retina$.
We adopt
Dirichlet boundary conditions, requiring the vanishing of the potential on
the boundary, $\varphi(x,t)=0$ on $\partial \Retina$, $\forall t$. Thus
 Eq.~\eqref{eq:dumped-wave} becomes
\begin{equation}
\begin{cases}
\mass\varphi_{tt}(x,t)+\drag \varphi_t(x,t) = \Delta \varphi(x,t)
+ \mu(x,t)& \hbox{in}\quad \Retina\times(0,+\infty);\\
\varphi(x,t)=0& \hbox{in}\quad \partial\Retina\times(0,+\infty);\\
\varphi(x,0)=0,\quad \varphi_t(x,0)=0& \hbox{in}\quad \Retina\times\{t=0\},
\end{cases}
\end{equation}
where from now on we decided to set $c=1$.
The pure wave equation corresponds to $\mass=1$ and $\drag = 0$, while with
$\mass =1$ and $\drag \neq 0$ we get a damped wave model.
Spurious reflections originating from the boundary are avoided through this
dumping term since, through an appropriate choice of the parameters, out-going waves are suppressed before they
can reach the boundary. The case of the heat equation is recovered with $\mass
=0$ with a diffusion coefficient equal to $1/\drag$.

The first step to numerically solve the update equation for the potential is
to discretize both the retina, considering a mesh $M=\{(i,j)\in\R^2 :
i=0,\dots h-1,\quad j=0,\dots,w-1\}$
of $h \times w$ points
(pixels), and the time interval with steps of length $\tau$. In the case
where we process visual streams, $\tau$ is synced with the temporal
resolution of the video by choosing it to be the inverse of the frame
rate of the stream. For static images, $\tau$ is a priori
fixed to an arbitrary
value. Then, we adopt the so called \textit{finite difference method},
approximating spatial and temporal derivatives through finite
differences. Considering an arbitrary pixel $(i,j)$ of the retina at a
certain time $t$, the evaluation of spatial (temporal) derivatives of the
potential in this point will just require the knowledge of the potential in
its adjacent points in space (time). According to the chosen approximations
for the derivatives, the unknown value of the potential at the following time
step is determined by a set of algebraic equations (\textit{explicit
methods}) or by a set of coupled equations (\textit{implicit
methods}). Generally speaking, implicit methods are less afflicted by numerical
instabilities, even though they are much slower than the
explicit ones~\cite{langtangen2017}. It should be noticed that in our implementation the potential is rescaled by a costant multiplicative factor, i.e. $\varphi \to k \varphi$. We performed a grid search to select the best hyper-parameter $k$. We found out that $k= 250000$  maximized the performance in both saliency and scanpath prediction tasks.

In our experiments, we have chosen backward finite difference approximations
for the time derivatives and a central one for the spatial derivatives,
\[
\begin{split}
&\left.\partial_{tt}\varphi(x,t)\right|^{t_{n+1}}_{i,j}\simeq \frac{\varphi_{i,j}^{t_{n+1}}-2\varphi_{i,j}^{t_n}+\varphi_{i,j}^{t_{n-1}}}{\tau^2},  \qquad\qquad \left.\partial_t \varphi(x,t)\right|^{t_{n+1}}_{i,j}\simeq \frac{\varphi_{i,j}^{t_{n+1}}-\varphi_{i,j}^{t_n}}{\tau}, \\
&\left.\partial_{xx}\varphi(x,t)\right|^{t_n}_{i,j}\simeq \varphi_{i,j+1}^{t_n}-2\varphi_{i,j}^{t_n}+\varphi_{i,j-1}^{t_n},\qquad\qquad
\left.\partial_{yy}\varphi(x,t)\right|^{t_n}_{i,j}\simeq\varphi_{i+1,j}^{t_n}-2\varphi_{i,j}^{t_n}+\varphi_{i-1,j}^{t_n},
\end{split}
\]
resulting in a implicit scheme.
To test our model, we have chosen two different sets of parameters,
summarized in Tab.~\ref{tab:tested_models}. The $\h$ model corresponds to the
pure diffusion case, while the $\dw$ model also considers a non vanishing
$\mass$ term.

\begin{wraptable}{r}{7.5cm}
\centering
\caption{Parameters of the $\h$ and $\dw$ models}
\label{tab:tested_models}
\medskip
\begin{tabular}{lcccccc} 
\toprule
\textbf{Model}&$\mass$&$\drag$&$c^2$&$\alpha_1$&$\alpha_2$&$\lambda$\\
\midrule
$\h$&0&$1/2500$&1&1&1&5\\
$\dw$&$1/25000$&$1/100$&1&1&1&5\\
\bottomrule
\end{tabular}
\end{wraptable}

\paragraph{Qualitative results}
In Fig.~\ref{fig:images} and Fig.~\ref{fig:videos} some qualitative results are
reported. Fig.~\ref{fig:images} shows some sequences of fixations obtained
through our model. The input static images come from the CAT2000 training
set, with an exposition time of 5 sec. In addition, Fig.~\ref{fig:videos} shows the dynamical evolution of the potential, in a $5$ sec. exploration of a static image. From this qualitative analysis we can underline
that the model $\h$ seems to have a stronger bias towards the centre of the
visual scene than model $\dw$. Moreover, model $\h$
seems to heavily smooth out the
microscopic structure of the mass density distribution, making the model $\dw$
probably preferable despite its slightly worst results in the
measures reported in the next sections.

\input figures_new.tex

\subsection{Saliency prediction}
\input saliency_prediction.tex

\subsection{Scanpath prediction}
\input scanpath_prediction.tex

\begin{table}[t]
\centering
\caption{\textbf{Saliency prediction scores.} AUC/NSS are similarity
metrics. Larger values are preferable.
\textbf{Scanpath prediction scores.} STDE is a similarity metric
while SED is a distance. Larger STDE scores are preferable, while smaller SED
score correspond with better results.
Best in bold.}
\label{tab:saliency-scanpathscores}
\medskip
\begin{tabular}{lccccccccc}
\toprule
 & &\qquad&\multicolumn{2}{c}{\textbf{Saliency Pred.}}&\qquad&\multicolumn{4}{c}{\textbf{Scanpath Pred.}}\\
 & &\qquad&AUC&NSS&\qquad&\multicolumn{2}{c}{SED}&\multicolumn{2}{c}{STDE}\\
\textbf{Model}&\textbf{Supervised}&\qquad& & &\qquad& \textit{mean}& \textit{best}&  \textit{mean}& \textit{best}\\
\midrule
GRAV~\cite{zanca2019gravitational}&No&\qquad   &0.84&1.57     &\quad&\textbf{7.34}&\textbf{3.72}&0.81&0.85\\
Eymol~\cite{zanca2017variational} &No&\qquad&  0.83&1.784     &\quad&7.94&4.10&0.74&0.81\\                          
SAM~\cite{sam} &Yes&\qquad&\textbf{0.88}&\textbf{2.38}        &\quad&8.02&4.25&0.77&0.83\\         
Deep Gaze II~\cite{kummerer2016deepgaze}&Yes&\qquad&0.77&1.16 &\quad&8.17&4.34&0.72&0.79\\         
Itti~\cite{itti1998model} &No&\qquad&0.77&1.06                &\quad&8.15&4.36&0.70&0.76\\         
\midrule                                                                              
{Our $\h$} &No&\qquad&0.84&1.69                        &\quad&7.73&3.85&\textbf{0.87}&\textbf{0.90}\\
{Our $\dw$} &No&\qquad&0.84&1.56                       &\quad&7.69&3.88&0.86&\textbf{0.90}\\
\bottomrule
\end{tabular}
\end{table}


%% file: figures_new.tex
\begin{figure}[htb]
    \centering 
\begin{subfigure}{0.33\textwidth}
  \includegraphics[width=\linewidth]{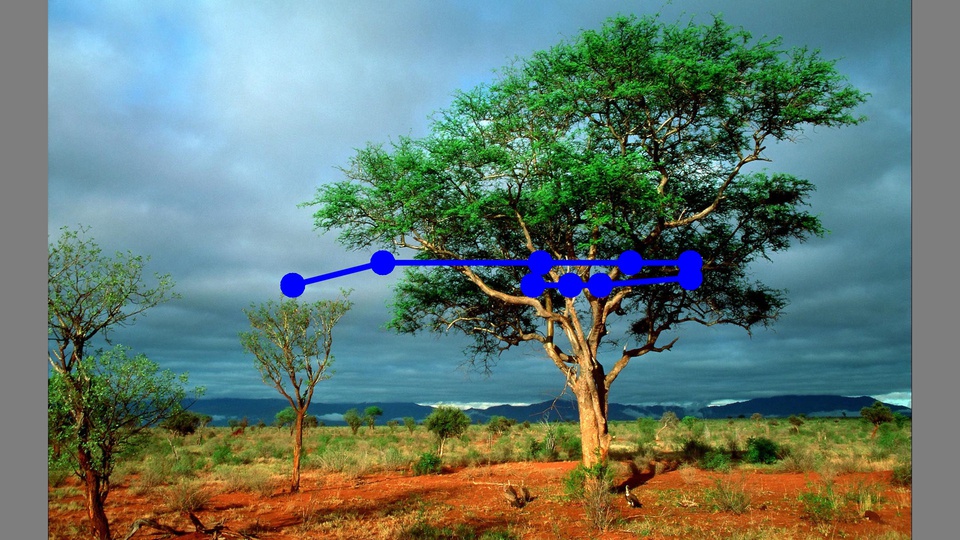}
\end{subfigure}\hfil 
\begin{subfigure}{0.33\textwidth}
  \includegraphics[width=\linewidth]{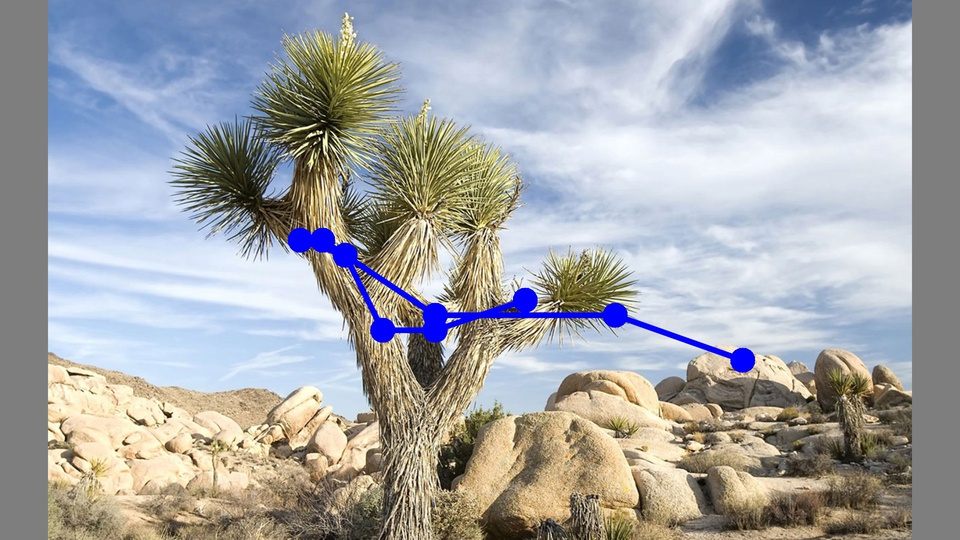}
\end{subfigure}\hfil 
\begin{subfigure}{0.33\textwidth}
  \includegraphics[width=\linewidth]{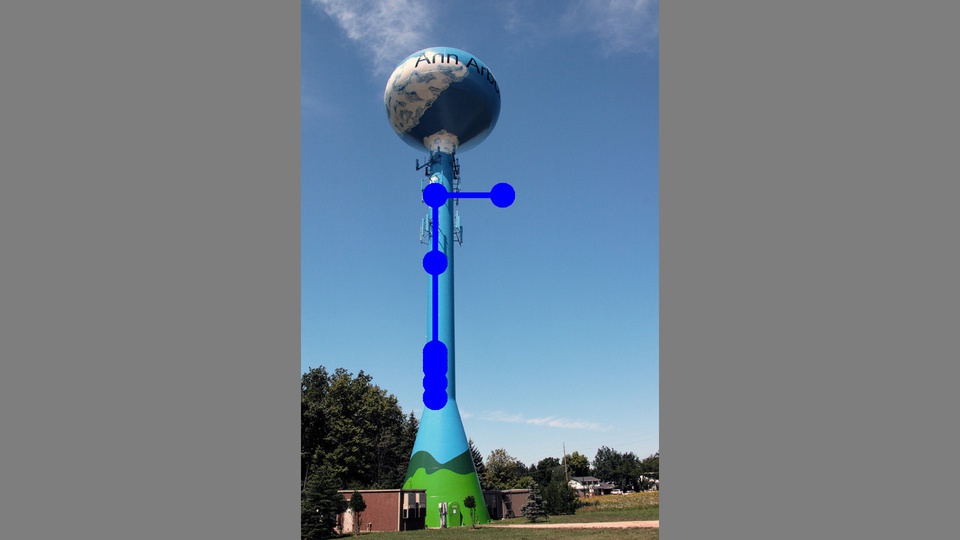}
\end{subfigure}

\vspace{1pt}

\begin{subfigure}{0.33\textwidth}
  \includegraphics[width=\linewidth]{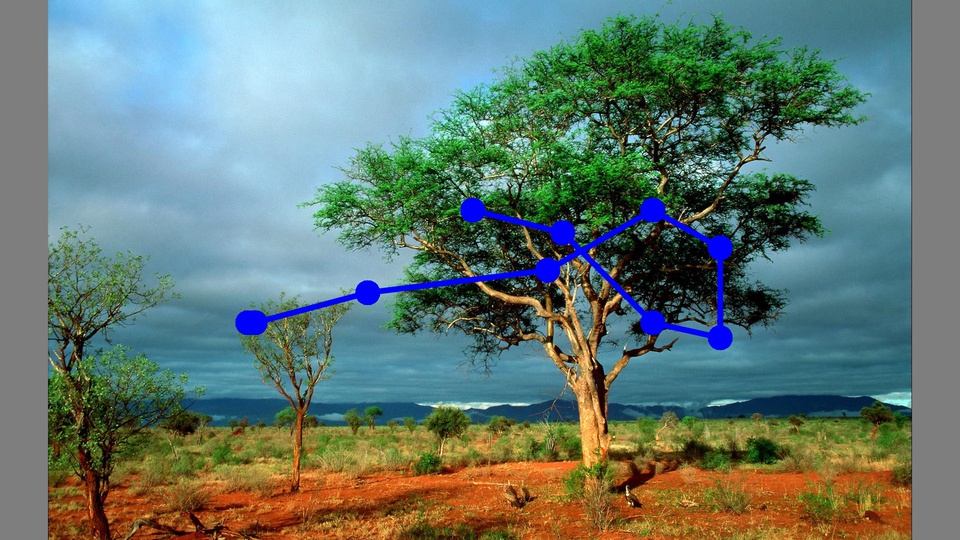}
\end{subfigure}\hfil 
\begin{subfigure}{0.33\textwidth}
  \includegraphics[width=\linewidth]{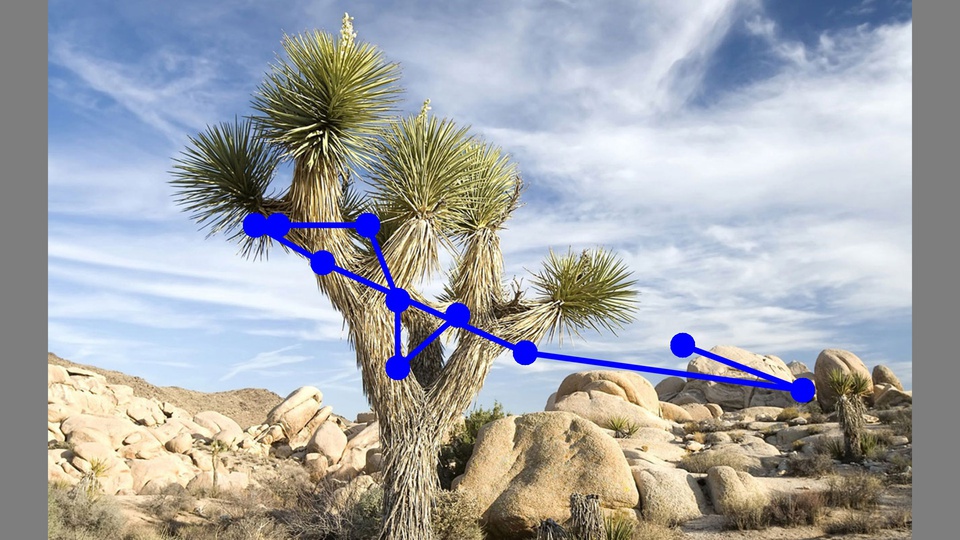}
\end{subfigure}\hfil 
\begin{subfigure}{0.33\textwidth}
  \includegraphics[width=\linewidth]{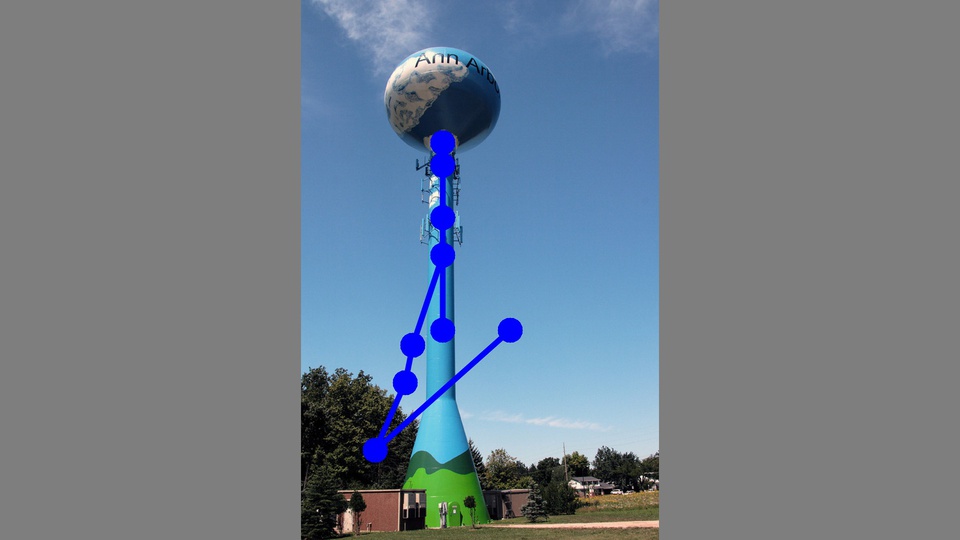}
\end{subfigure}

\vspace{5pt}

\begin{subfigure}{0.33\textwidth}
  \includegraphics[width=\linewidth]{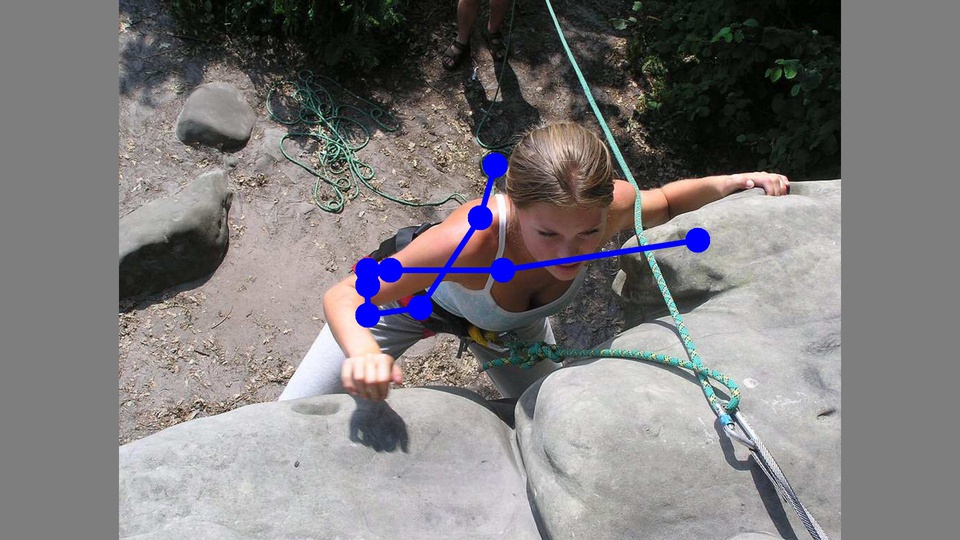}
\end{subfigure}\hfil 
\begin{subfigure}{0.33\textwidth}
  \includegraphics[width=\linewidth]{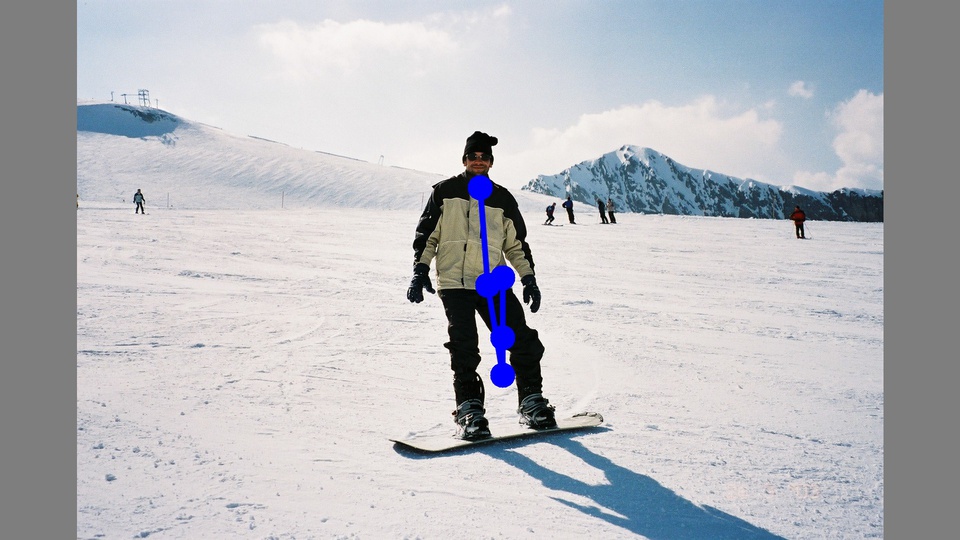}
\end{subfigure}\hfil 
\begin{subfigure}{0.33\textwidth}
  \includegraphics[width=\linewidth]{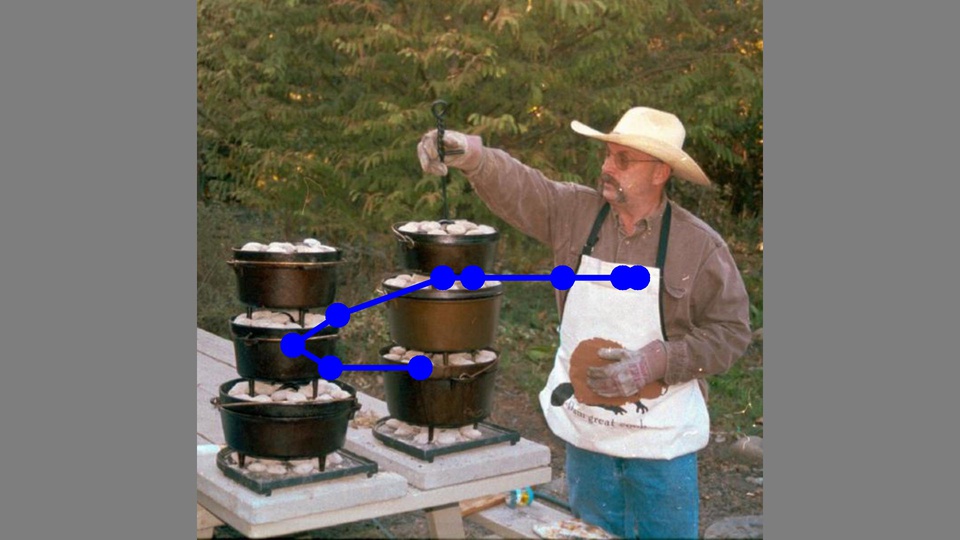}
\end{subfigure}

\vspace{1pt}
\begin{subfigure}{0.33\textwidth}
  \includegraphics[width=\linewidth]{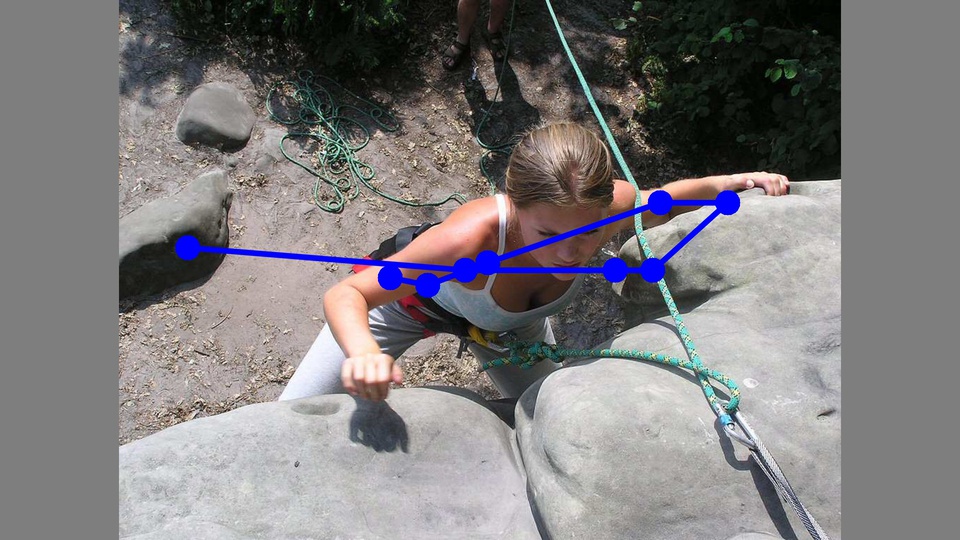}
\end{subfigure}\hfil 
\begin{subfigure}{0.33\textwidth}
  \includegraphics[width=\linewidth]{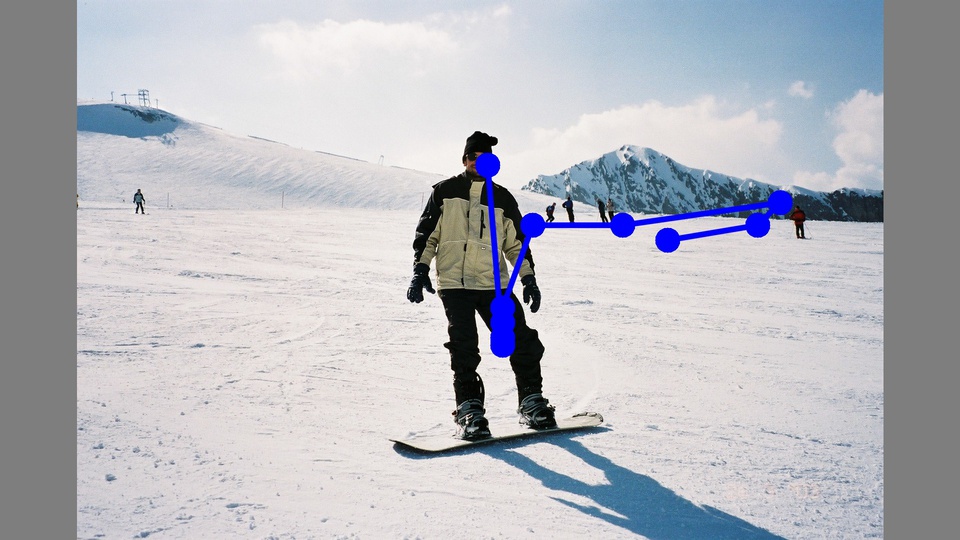}
\end{subfigure}\hfil 
\begin{subfigure}{0.33\textwidth}
  \includegraphics[width=\linewidth]{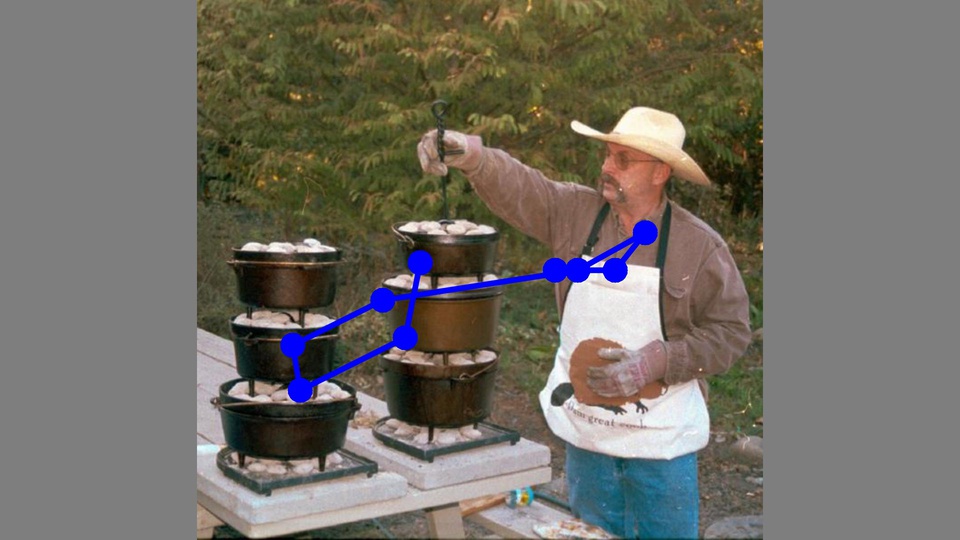}
\end{subfigure}

\caption{Fixations for the $\h$ model (top) and $\dw$ model (bottom)
obtained in five seconds of exploration. Images from the CAT2000 training set.}
\label{fig:images}
\end{figure}

\begin{figure}[htb]
    \centering 
\begin{subfigure}{0.45\textwidth}
  \includegraphics[width=\linewidth]{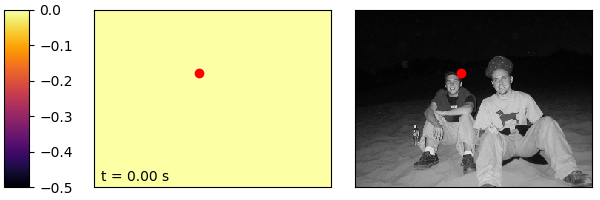}
\end{subfigure}\hfil 
\begin{subfigure}{0.45\textwidth}
  \includegraphics[width=\linewidth]{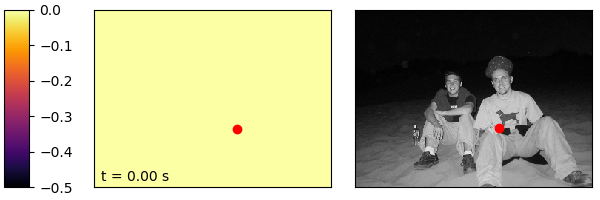}
\end{subfigure}

\vspace{1pt}

\begin{subfigure}{0.45\textwidth}
  \includegraphics[width=\linewidth]{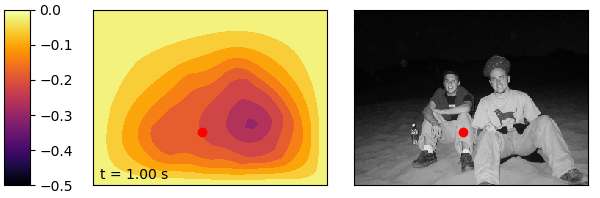}
\end{subfigure}\hfil 
\begin{subfigure}{0.45\textwidth}
  \includegraphics[width=\linewidth]{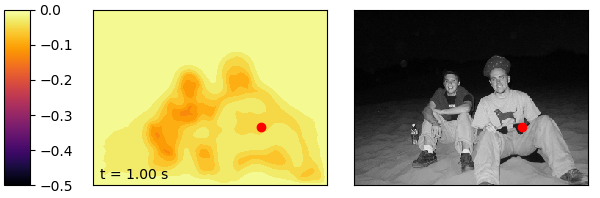}
\end{subfigure}

\vspace{1pt}
\begin{subfigure}{0.45\textwidth}
  \includegraphics[width=\linewidth]{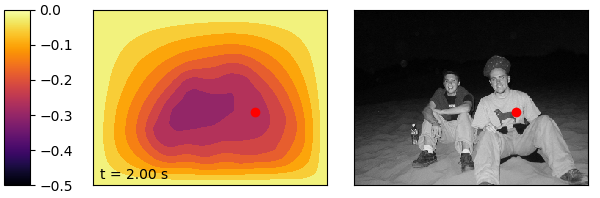}
\end{subfigure}\hfil 
\begin{subfigure}{0.45\textwidth}
  \includegraphics[width=\linewidth]{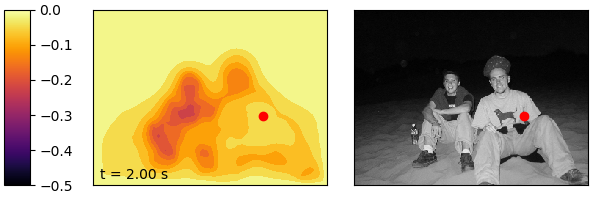}
\end{subfigure}

\vspace{1pt}
\begin{subfigure}{0.45\textwidth}
  \includegraphics[width=\linewidth]{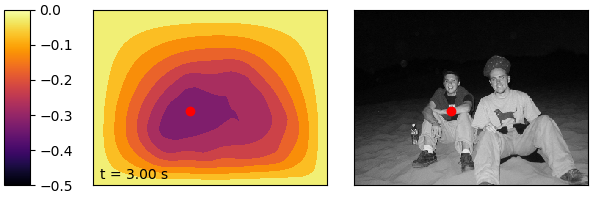}
\end{subfigure}\hfil 
\begin{subfigure}{0.45\textwidth}
  \includegraphics[width=\linewidth]{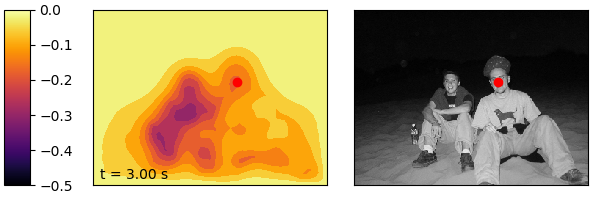}
\end{subfigure}

\vspace{1pt}
\begin{subfigure}{0.45\textwidth}
  \includegraphics[width=\linewidth]{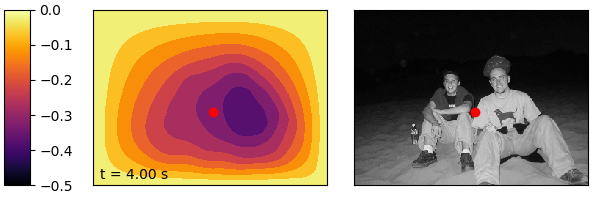}
\end{subfigure}\hfil 
\begin{subfigure}{0.45\textwidth}
  \includegraphics[width=\linewidth]{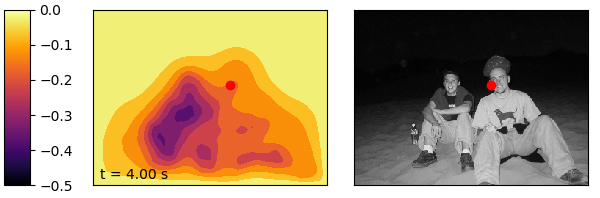}
\end{subfigure}

\vspace{1pt}
\begin{subfigure}{0.45\textwidth}
  \includegraphics[width=\linewidth]{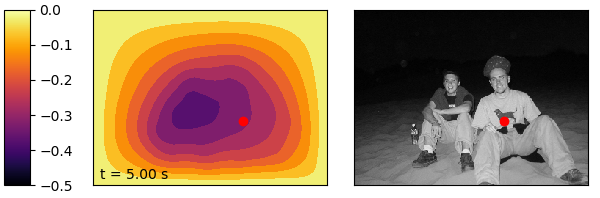}
\end{subfigure}\hfil 
\begin{subfigure}{0.45\textwidth}
  \includegraphics[width=\linewidth]{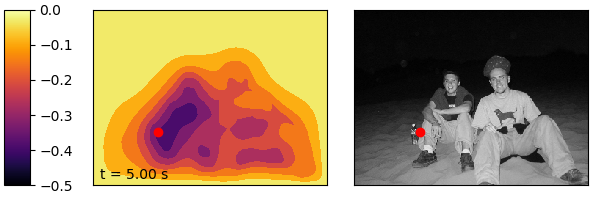}
\end{subfigure}

\caption{Dynamical evolution of the potential in the $\h$ model (left)
and $\dw$ model (right) in $5$ seconds of exploration of a static image.}
\label{fig:videos}
\end{figure}

%% file: saliency_prediction.tex
The saliency prediction task consists of generating saliency maps to predict
the probability of each pixel to be attended by a human subject during
free-viewing~\cite{borji2013analysis}. We exploit our model to generate
simulated scanpaths by means of numerical integration, as described in the previous section. Different visual explorations are obtained by
initializing the system with arbitrary different initial conditions. Since
human subjects during free-viewing are usually asked to look at a target
point in the center of the screen before a visual stimulus is presented, we
choose to initialize the model within a range of $2.5^\circ$ of visual angle
in the center of the image and with focus of attention velocity close to
zero. Fixations extracted from the simulated scanpath were accumulated in the
so-called fixation map. Finally, a saliency map was obtained by applying a
Gaussian smoothing to the fixation map. It is well
known~\cite{salMetrics_Bylinskii,borji2012state,kummerer2015information} that
when applying a certain amount of smoothing and center bias, 
we can improve the performance of models in saliency prediction. 
Therefore, we performed a grid
search on blurring and center bias parameters to determine the best estimate
associated with the model. Metrics exist to compare the generated saliency
map with the human one. Here we compare the performance of our model with
respect to two different saliency metrics:
\begin{itemize}
\item \textit{Area Under the ROC Curve~\cite{Judd} (AUC)}. The saliency map
is treated as a binary classifier to separate positive from negative samples
at various thresholds. The true positive rate is the proportion of saliency
map values above threshold at fixation locations. The false positive rate is
the proportion of saliency map values above threshold at all pixels. 
\item \textit{Normalized Scanpath Saliency~\cite{peters2005components}
(NSS)}. It measures the mean saliency value at fixated locations of the
saliency map, normalized with zero mean and unit variance.
\end{itemize}
Results are summarized in Tab.~\ref{tab:saliency-scanpathscores}. The model is
compared with five different models. Two of
these~\cite{kummerer2016deepgaze,sam} are state-of-the-art
supervised\footnote{We indicate as supervised those approaches in which saliency is learned directly from data by machine learning techniques.}
saliency models. Two
others~\cite{zanca2019gravitational,zanca2017variational} are
state-of-the-art unsupervised scanpath models. The last one is the classic
Itti's model~\cite{itti1998model}. All models were used in their original
software implementation provided by the respective authors. We notice that supervised models
maintain the state of the art in the estimation of saliency. This may be due to the fact that
they can learn semantic properties of the image, which highly correlate with human visual
attention\cite{theeuwes2000time,connor2004visual}. However, the proposed models achieve state of the art results among
unsupervised models and competes very well with supervised models.


%% file: scanpath_prediction.tex
The scanpath prediction task consists of predicting the sequence of fixations
that a human subject performs in free-viewing condition when presented with a
stimulus. Unlike the saliency prediction task, here the temporal dynamics of the attention mechanism are evaluated. The models were asked to predict \textit{where} and \textit{in which order} the subject will perform fixations on the scene. We used two different evaluation metrics:
\begin{itemize}
\item \textit{String-edit distance~\cite{Foulsham} (SED)}. The input stimulus
is divided into $n \times n$ regions, labeled with characters. Scanpaths is
turned into strings by associating each fixation with the corresponding
region. Finally, the string-edit algorithm is used to provide a measure of
the distance between the two generated strings.
\item \textit{Scaled time-delay embeddings~\cite{zanca2020toward}
(STDE)}. This measure derives from quantitative methods in physics to compare
stochastic and dynamic scanpaths of varied lengths.  It is defined as the
average of the minimum Euclidean distances of each sub-sequence of length $n$
from a target trajectory. Coordinates are normalized between zero and one to
obtain comparable measure for images of different sizes.
\end{itemize} 
Difficulties arise when evaluating visual attention models in tasks of
scanpath prediction~\cite{zanca2020toward}. 
We show the results in terms of \textit{mean} and
\textit{best} prediction scores. In the case of \textit{mean}, scores are
averaged over all subjects in the dataset; in the case of \textit{best}, we
consider, for each of the simulated scanpaths, only the subject that best 
matches with that simulated scanpath.

The proposed models are
compared with the same five models considered in the previous
experiment. Whenever possible, we used the authors' original implementation to
generate fixation sequences. For strictly saliency-oriented models, we have
applied the winner-take-all algorithm~\cite{koch1987shifts}. 
We selected the location with the highest saliency value, 
then we inhibited in a radius of 4 degrees~\cite{hooge2000inhibition} of visual angle 
and selected the next fixation. The results are summarized in
Tab.~\ref{tab:saliency-scanpathscores}. They show that the
proposed model reaches the state of the art in scanpath prediction with
respect to  the STDE metric. Better results in terms of STDE indicate a greater 
adherence to the shape of the target human trajectories, 
while the discretized spatial grid makes the
SED less precise to evaluate spatial properties. The advantage of our model
over the other approaches is likely due to the fact that the local
implementation induces a proximity preference~\cite{koch1987shifts} in the choice of the following target. 
Supervised models that benefit from
the possibility of learning semantic
characteristics of the scenes in a data-driven approach, are instead outclassed by unsupervised models
in this scanpath prediction task.


%% file: conclusions.tex
In this paper, we have presented a  computational model
of focus of attention that is  inspired from classic wave and diffusion 
propagation mechanisms, that are  joined with the principle
of inhibition of return.  It is proven that the resulting field gets
arbitrarily close to recently proposed gravitational-based models of focus of
attention models, which explains the reasons of the state of the art
experimental results achieved in scanpath prediction tasks. 
The connection with gravitational model arises because 
of a fundamental regularization 
property that is proven to reduce the proposed model to Poisson's
equation. The proposed theory clearly explains the emergence of
reactions to peripheral visual stimuli.
In particular, a distinctive feature of the proposed computational model 
is its spatiotemporal locality, which makes it very well-suited for SIMD parallel 
implementations, and also represents a fundamental property 
typically required for biological plausibility. 

While the current numerical discretization is based on the
traditional 2D gridding of the retina, one can think of approximating the
continuous differential equations by higher-order approximations 
of spatial derivatives, which improves the approximation 
and offers additional insights on the biological plausibility of the model.

%% file: supp.tex
\setcounter{page}{1}
\setcounter{equation}{0}
\centerline{\Large\bf Supplementary Material}
\bigskip
\meno
{\bf Notation. } We will indicate with $B_r(x)$ the
ball of radius $r$ and center $x$, with $S_r(x)$ the
sphere with the same radius and  center. We will
furthermore indicate with $d\H^{n-1}$ the $n-1$ dimensional
Hausdorff measure in $\R^n$ (for surface integrals).
Let $\Omega$ be an open set in $\R^n$ and $D$ a domain
with $\overline D\subset\Omega$.
For any regular  function
$f\colon \overline D\to \R$ we define 
\[
\mint_{D}f(x)\, dx:=\frac{1}{m_n(D)}\int_{D}f(x)\, dx,\qquad
m_n(D):=\int_D 1\, dx.
\]
The supplementary material is organized as follows: In~\ref{Appendix-Duhamel}
we give a brief summary of Duhamel's Principle (a good reference for this
topic is~\cite{evans2010partial}) then in Section~\ref{proof:th1}
we give a proof of Theorem~1 from the main article.

\section{Duhamel's Principle}
\label{Appendix-Duhamel}
Consider the following abstract form of
the Cauchy problem for an evolution equation of the first order:
\[P:\,\left\{
\begin{aligned}
&u'(t)= A u(t)+f(t)\qquad (t>0);\\
&u(0)=\varphi,
\end{aligned}\right.\]
where the spatial dependece of $u$ is not explicitly written
and $A$ is a differential operator with respect to the spatial
variables. Now consider for every
fixed $s\in [0,t]$ the additional homogeneous problem
\[P':\, \left\{
\begin{aligned}
&v'(t)= A v(t)\qquad (t>0);\\
&v(0)=f(s)+ A\varphi,
\end{aligned}\right.\]
and let us indicate with $v_s(t)$ the solution to this problem.
Then we can directly check that the function
\[u(t)=\varphi +\int_0^t v_s(t-s)\, ds,\]
solves problem $P$. Indeed we have $u(0)=0$ and
\[u'(t)= v_t(0)+\int_0^t v_s'(t-s)\, ds=f(t)+A u(t).\]
In a similar way we can treat the second order problem
\[Q:\, \left\{
\begin{aligned}
&u''(t)= A u(t)+f(t)\qquad (t>0);\cr
&u(0)=\varphi;\\
&u'(0)=\psi.\cr
\end{aligned}\right.\]
This time for any $s\in[0,t]$ we consider the solution
$v_s$ to the problem
\[Q':\,\left\{
\begin{aligned}
&v''(t)= A u(t)\qquad (t>0);\cr
&v(0)=0;\cr
&v'(0)=f(s)+A\varphi+s A\psi.\cr
\end{aligned}\right.\]
Then one can verify that the function
\[u(t)=\varphi+t\psi+\int_0^tv_s(t-s)\, ds\]
is a solution to problem $Q$.

\section{Proof of Theorem~1}\label{proof:th1}
We prove the claim first for the solutions to the heat equation
$\h$ and then for the solution to the wave equations $\w$.

\paragraph{\bf Heat equation}
In order to find an explicit formula for the solution to the problem
\begin{equation}\label{heat-supp}
\begin{cases}
\varphi_t=c(\laplace\varphi+\mu)& \hbox{in}\quad \R^2\times(0,+\infty);\\
\varphi(x,0)=0,& \hbox{in}\quad \R^2\times\{t=0\},
\end{cases}
\end{equation}
we begin considering the problem
\begin{equation}
\begin{cases}
u_t=c\laplace u& \hbox{in}\quad \R^2\times(0,+\infty);\\
u(x,0)=u_0(x),& \hbox{in}\quad \R^2\times\{t=0\}.
\end{cases}
\label{c-heat}
\end{equation}
For this particular PDE the constant $c$ can be absorbed entirely by a
time rescaling $t\to ct$. So that the solution $u$  to
Eq.~\eqref{c-heat} can be written as $u(x,t)=v(x,ct)$ where $v$ solves
\begin{equation}
\begin{cases}
v_t=\laplace v& \hbox{in}\quad \R^2\times(0,+\infty);\\
v(x,0)=u_0(x),& \hbox{in}\quad \R^2\times\{t=0\}.
\end{cases}
\label{plain-heat}
\end{equation}
Finally the solution to this problem can be found in terms of the
well known heat kernel
\begin{equation}
U(x,t):=\frac{1}{4\pi t} e^{-|x|^2/4t},
\end{equation}
as $v(x,t)=(U(\cdot,t)*u_0)(x)$. Then the  solution to~\eqref{c-heat} is
\[
u(x,t)=\frac{1}{4\pi ct}\int_{\R^2} e^{-|x-y|^2/4ct}u_0(y)\, dy.
\]
Applying Duhamel's principle to Eq.~\eqref{heat-supp} yields 
\[
\varphi(x,t)=\int_0^t\frac{1}{4\pi(t-s)}\int_{\R^2}
e^{-|x-y|^2/4c(t-s)}\mu(y,s)\, dyds.
\]
Now let $c(t-s)=\tau$, then
\[
\varphi(x,t)=\frac{1}{4\pi}
\int_0^{ct}\int_{\R^2}
\frac{e^{-|x-y|^2/4\tau}}{\tau}\mu(y,t-\tau/c)\, dyd\tau.
\]
The gradient of $\varphi$ can now be directly calculated as
\[
\nabla\varphi(x,t)=-\frac{1}{8\pi}
\int_0^{ct}\int_{\R^2}
\frac{e^{-|x-y|^2/4\tau}}{\tau^2}(x-y)\mu(y,t-\tau/c)\, dyd\tau.
\]

Taking the formal limit $c\to+\infty$ we obtain
\[
\nabla\varphi(x,t)\to-\frac{1}{8\pi}
\int_{\R^2}
\biggl(\int_0^{+\infty}\frac{e^{-|x-y|^2/4\tau}}{\tau^2}\, d\tau\biggr)
(x-y)\mu(y,t) \, dy.
\]
Because $\int_0^\infty e^{-a^2/\tau}/\tau^2\, d\tau=a^{-2}$, we have
\[
\nabla\varphi(x,t)\to-\frac{1}{2\pi}
\int_{\R^2}\frac{x-y}{|x-y|^2}\mu(y,t) \, dy,
\]
which is indeed the gradient of the potential that solves Poisson equation
with source $\mu$. Notice however that performing the formal limit
$c\to\infty$ directly into the expression for the potential would lead to
a divergent limit due to the divergent quantity
$\int_0^\infty e^{-a^2/\tau}/\tau\,
d\tau$.

\draftA{In this case the limit $c\to\infty$ correspond to the asymptotic
value of $v$. So it can be given a gradient flow interpretation...}

\paragraph{Wave equation} Consider now the 
nonhomogeneous wave eqaution
\begin{equation}
\begin{cases}
\varphi_{tt}=c^2(\laplace\varphi+\mu)& \hbox{in}\quad \R^2\times(0,+\infty);\\
\varphi(x,0)=0,\quad \varphi_t(x,0)=0& \hbox{in}\quad \R^2\times\{t=0\}.
\end{cases}\label{nonhom-wave}
\end{equation}
In order to find an explicit solution to this problem, as usual
(see \cite{evans2010partial}), we start by considering the related problem
\begin{equation}
\begin{cases}
u_{tt}-c^2\laplace u=0& \hbox{in}\quad \R^2\times(0,+\infty);\\
u(x,0)=f(x),\quad u_t(x,0)=g(x)& \hbox{in}\quad \R^2\times\{t=0\}.
\end{cases}
\label{c-wave}
\end{equation}
The solution of such problem is $u(x,t)=v(x,ct)$, where $v$ solves
\begin{equation}
\begin{cases}
v_{tt}-\laplace v=0& \hbox{in}\quad \R^2\times(0,+\infty);\\
v(x,0)=f(x),\quad v_t(x,0)=g(x)/c& \hbox{in}\quad \R^2\times\{0\}.
\end{cases}
\label{plain-wave}
\end{equation}
Since the solution of~\eqref{plain-wave} is given by the Poisson's formula
in two dimensions
\[
v(x,t)={1\over 2}\mint_{B_t(x)}\frac{t f(y)+c^{-1}t^2g(y)+
t\nabla f(y)\cdot (y-x)}
{(t^2-\vert y-x\vert^2)^{1/2}}\, dy,
\]
we have
\begin{equation}
u(x,t)={1\over 2}\mint_{B_{ct}(x)}\frac{ct f(y)+ ct^2g(y)+
ct\nabla f(y)\cdot (y-x)}
{(c^2t^2-\vert y-x\vert^2)^{1/2}}\, dy.
\label{c-wave-sol}
\end{equation}
The solution to Eq.~\eqref{nonhom-wave} can be obtained from the solution
of Eq~\eqref{plain-wave} via the Duhamel's principle (see
Section~\ref{Appendix-Duhamel}). In this case we have that
\begin{equation}\varphi(x,t)=\int_0^t w_s(x,t-s)\, ds,
\label{Duhamel-wave}\end{equation}
where $w_s$ solves
\[
\begin{cases}
w_{tt}-c^2\laplace w=0& \hbox{in}\quad \R^2\times(0,+\infty);\\
w(x,0)=0,\quad w_t(x,0)=c^2\mu(x,s)& \hbox{in}\quad \R^2\times\{0\}.
\end{cases}
\]
Thus from Eq.~\eqref{c-wave-sol} and~\eqref{Duhamel-wave} we immediately have:
\[
\varphi(x,t)=\frac{1}{2\pi}\int_0^t
\int_{B_{c(t-s)}(x)}\frac{c\mu(y,s)}
{(c^2(t-s)^2-\vert y-x\vert^2)^{1/2}}\, dyds.
\]
Now let us make the change of variables $c(t-s)=\tau$ in the integral over
$s$; we thus obtain:
\begin{equation}
\varphi(x,t)=\frac{1}{2\pi}\int_0^{ct}
\int_{B_{\tau}(x)}\frac{\mu(y,t-\tau/c)}
{(\tau^2-\vert y-x\vert^2)^{1/2}}\, dy d\tau.
\label{wave-pot-1-form}
\end{equation}

\begin{example}
In order to understand Eq.~\eqref{wave-pot-1-form} let us consider
the case of a unit mass fixed at the origin: $\mu(x,t)=\delta_x$.
In this case
\begin{equation}
\begin{aligned}
\varphi(x,t)=\frac{1}{2\pi}\int_0^{ct}
\int_{B_{\tau}(x)}\frac{\delta_y}
{(\tau^2-\vert y-x\vert^2)^{1/2}}\, dy d\tau
&=\frac{1}{2\pi}\int_{|x|}^{ct}
\frac{1}
{(\tau^2-\vert x\vert^2)^{1/2}}\, d\tau\\
&=\frac{1}{2\pi}\left[\log\bigl|\tau+\sqrt{\tau^2-|x|^2}\,\bigr|\right]_{|x|}^{ct}\\
&=\frac{1}{2\pi}\log\bigl(ct+\sqrt{(ct)^2-|x|^2}\,\bigr)
+\frac{1}{2\pi}\log\frac{1}{|x|}.
\end{aligned}
\end{equation}
Then
\begin{equation}
\varphi(x,t)=\frac{1}{2\pi}\log(ct)
+\frac{1}{2\pi}\log\bigl(1+\sqrt{1-(|x|/ct)^2}\,\bigr)
+\frac{1}{2\pi}\log\frac{1}{|x|}.
\end{equation}
Notice that in the last formula as $c\to\infty$ we have a divergent part plus
a finite part which is indeed our initial guess for this limit;
moreover the divergent part has a vanishing spatial gradient meaning that
it does not effect the force which is entirely given by gradient of
$\log(1/|x|)$ which is indeed the force that ones derive from Poisson
equation on $\R^2$.
\end{example}

This example suggests to look for the convergence of $\nabla\varphi$ rather
than that of $\varphi$ itself since the latter
quantity  can give rise to divergences.

In general let us now come back to Eq.~\eqref{wave-pot-1-form}. This integral
is performed over a cone in the space $y$-$\tau$. A little thinking shows
that such integration can be rearranged as follows:
\begin{equation}
\varphi(x,t)=\frac{1}{2\pi}\int_{B_{ct}(x)}\biggl(\int_{|y-x|}^{ct}
\frac{\mu(y,t-\tau/c)}
{(\tau^2-\vert y-x\vert^2)^{1/2}}\,d\tau\biggr)\,dy .
\label{wave-pot-2-form}
\end{equation}
Or, equivalently, performing the change of variables $z=y-x$
\begin{equation}
\varphi(x,t)=\frac{1}{2\pi}\int_{B_{ct}(0)}\biggl(\int_{|z|}^{ct}
\frac{\mu(z+x,t-\tau/c)}
{(\tau^2-\vert z\vert^2)^{1/2}}\,d\tau\biggr)\,dz .
\end{equation}
Since we are interested in the limit $c\to\infty$ we can expand
$\mu(z+x,t-\tau/c)$ in powers of $1/c$ around zero:
\begin{equation}
\mu(z+x,t-\tau/c)= \mu(z+x,t)-\mu_t(z+x,t)\tau\frac{1}{c}+
\mu_{tt}(z+x,t)\tau^2\frac{1}{c^2}+o(1/c^2).
\label{expansion}
\end{equation}
At order zero in $1/c$ we have
\begin{equation}
\begin{aligned}
\varphi(x,t)&=\frac{1}{2\pi}\int_{B_{ct}(0)}\biggl(\int_{|z|}^{ct}
\frac{1}
{(\tau^2-\vert z\vert^2)^{1/2}}\,d\tau\biggr) \mu(z+x,t)\,dz\\
&=\frac{1}{2\pi}\int_{B_{ct}(0)}\biggl(
\log(ct)+\log\bigl(1+\sqrt{1-|z^2|/(ct)^2}\,\bigr)+\log\frac{1}{|z|}\biggr)
\mu(z+x,t)\,dz.
\end{aligned}
\end{equation}
The gradient of such expression is
\begin{equation}
\nabla\varphi(x,t)=\frac{1}{2\pi}\int_{B_{ct}(0)}\biggl(
\log(ct)+\log\bigl(1+\sqrt{1-|z^2|/(ct)^2}\,\bigr)+\log\frac{1}{|z|}\biggr)
\nabla\mu(z+x,t)\,dz.
\label{wave-pot-3-form}
\end{equation}
Now we can use the following version of the divergence theorem
\begin{equation}
\int_\Omega f\nabla g\, dx=\int_{\partial\Omega}fg \nu\, d\H^{n-1}
-\int_\Omega \nabla f g\, dx,
\label{divergence-formula}
\end{equation}
where $\nu$ is the normal to $\partial\Omega$. In order to prove this start from
the divergence theorem for vector fields:
\[\int_\Omega \div v\,dx=\int_{\partial\Omega} v\cdot \nu\, d\H^{n-1},\]
then choose $v_k=fg\delta_{ki}$, therefore 
\[
\int_\Omega f\nabla_i g\, dx
+\int_\Omega \nabla_i f g\, dx= \int_\Omega \div v\,dx
=\int_{\partial\Omega} fg\delta_{ki}\nu_k\, d\H^{n-1}
=\int_{\partial\Omega} fg\nu_i\, d\H^{n-1},
\]
which gives the wanted formula. If we apply such expression to
Eq.~\eqref{wave-pot-3-form} we get
\begin{equation}
\begin{aligned}
\nabla\varphi(x,t)=&\frac{1}{2\pi}\log(1)\int_{S_{ct}(0)}
\mu(z+x,t)\nu(z)\,d\H^1(z)\\
&\quad-\frac{1}{2\pi}\int_{B_{ct}(0)}\biggl(
\nabla\log\bigl(1+\sqrt{1-|z|^2/(ct)^2}\,\bigr)+\nabla\log\frac{1}{|z|}\biggr)
\mu(z+x,t)\,dz.
\end{aligned}
\end{equation}
Here we found by direct calculation (since $\log1=0$) that the boundary
term is null; notice that indeed this term must be zero also because
it comes from an integral evaluated betwee $|z|$ and $ct$,
so when $|z|=ct$ the whole term is vanishing.

Expanding the gradient in the remaining term we get
\begin{equation}
\nabla\varphi(x,t)=\frac{1}{2\pi}\int_{B_{ct}(0)}\biggl(
\frac{1}{c^2}\frac{z}{1-|z|^2/(ct)^2+\sqrt{1-|z|^2/(ct)^2}}
+\frac{z}{|z|^2}\biggr)
\mu(z+x,t)\,dz.
\end{equation}
As we take the formal limit  $c\to+\infty$, we obtain
\begin{equation}
\nabla\varphi(x,t)\to\frac{1}{2\pi}\int_{\R^2}
\frac{z}{|z|^2}\mu(z+x,t)\,dz=
-\frac{1}{2\pi}\int_{\R^2}
\frac{x-y}{|x-y|^2}\mu(y,t)\,dy,
\end{equation}
which is the wanted result.

We will now consider the generic term in~\eqref{expansion} to show that
we correctly performed the limit disregarding in the approximation of
$\mu(z+x,t-\tau/c)$ all the higher order terms in $1/c$.

Let us define
\[I^c_n(\xi):=\frac{1}{c^n}\int_{\xi}^{ct}
\frac{\tau^n}{(\tau^2-\xi^2)^{1/2}}\, d\tau.\]
Changing the integration variable  $s=\tau-\xi$ we
this quantity can be written as
\[I^c_n(\xi)=\frac{1}{c^n}\int_{0}^{ct-\xi}
\frac{(s+\xi)^n}{(s^2+2s\xi)^{1/2}}\, ds.\]
And its derivative (with respect to the parameter $\xi$) is 
\[
{I^c_n}'(\xi)=
-\frac{t^n}{\sqrt{(ct)^2-\xi^2}}+\frac{1}{c^n}\int_0^{ct-\xi}
\left(\frac{n(s+\xi)^{n-1}}{(s^2+2s\xi)^{1/2}}
-\frac{s(s+\xi)^n}{(s^2+2s\xi)^{3/2}}\right)\, ds
\]
In the last integral let us perform the additional change of variable $s=cr$
\[
{I^c_n}'(\xi)=
-\frac{t^n}{\sqrt{(ct)^2-\xi^2}}+\frac{1}{c}\int_0^{t-\xi/c}
\left(\frac{n(r+\xi/c)^{n-1}}{(r^2+2r\xi/c)^{1/2}}
-\frac{r(r+\xi/c)^n}{(r^2+2r\xi/c)^{3/2}}\right)\, dr
\]
As we formally let $c\to+\infty$ we have that the integral converges
to $(n-1)\int_0^t r^{n-2}$. For $n\ge2$ it is immediate to check
that
\[
{I^c_n}'(\xi)\to 0\quad \hbox{as $c\to+\infty$};
\]
for $n=1$ this property can be checked by direct calculations; indeed
\[
I_1^c(\xi)=\frac{1}{c}\sqrt{(ct)^2-\xi^2},\qquad {I^c_1}'(\xi)=
-\frac{1}{c}\frac{\xi}{\sqrt{(ct)^2-\xi^2}}.
\]
The quantity $I^c_n$ has been defined precisely so that
 the $n$-th term in the expansion in
powers of $1/c$ of the gradient of the potential would be
\begin{equation}
\frac{1}{2\pi}\int_{B_{ct}(0)}\biggl(\frac{1}{c^n}\int_{|z|}^{ct}
\frac{\tau^n}
{(\tau^2-\vert z\vert^2)^{1/2}}\,d\tau\biggr)
\nabla\partial_t^n\mu(z+x,t)\,dz=
\frac{1}{2\pi}\int_{B_{ct}(0)}I^c_n(|z|)
\nabla\partial_t^n\mu(z+x,t)\,dz.
\label{n-th-term}
\end{equation}
Since $I^c_n(ct)=0$ using~\eqref{divergence-formula} Eq.~\eqref{n-th-term}
becomes
\[
-\frac{1}{2\pi} \int_{B_{ct}(0)} {I^c_n}'(|z|)\frac{z}{|z|}
\partial_t^n\mu(z+x,t)\,dz,
\]
and this quantity for $n\ge1$ goes to $0$ as $c\to+\infty$. \draftA{Finally!!!}
